\documentclass[conference]{IEEEtran}
\IEEEoverridecommandlockouts
\usepackage{cite}
\usepackage{amsmath,amssymb,amsfonts}
\usepackage{algorithmic}
\usepackage{graphicx}
\usepackage{textcomp}
\usepackage{enumitem}
\usepackage{xcolor}
\usepackage{booktabs}
\usepackage{amsmath,amssymb,amsfonts}
\usepackage{caption}
\usepackage{subcaption}
\usepackage{bm}
\def\BibTeX{{\rm B\kern-.05em{\sc i\kern-.025em b}\kern-.08em
    T\kern-.1667em\lower.7ex\hbox{E}\kern-.125emX}}

\def\BibTeX{{\rm B\kern-.05em{\sc i\kern-.025em b}\kern-.08em
    T\kern-.1667em\lower.7ex\hbox{E}\kern-.125emX}}

\def\BibTeX{{\rm B\kern-.05em{\sc i\kern-.025em b}\kern-.08em
    T\kern-.1667em\lower.7ex\hbox{E}\kern-.125emX}}    
    
\begin{document}

% \title{Feasiblity Aanalysis of System Scenario-based Temperature Prediction for HVAC Management}
\title{Improving Building Temperature Forecasting: A Data-driven Approach with System Scenario Clustering
}

\author{\IEEEauthorblockN{Dafang Zhao\IEEEauthorrefmark{1}, Zheng Chen\IEEEauthorrefmark{2}, Zhengmao Li\IEEEauthorrefmark{3}, Xiaolei Yuan\IEEEauthorrefmark{4}, Ittetsu Taniguchi\IEEEauthorrefmark{1}}
  \IEEEauthorblockA{\IEEEauthorrefmark{1}Graduate School of Information Science and Technology, Osaka University, Japan\\\IEEEauthorrefmark{2} ISIR, Osaka University, Japan\\
  \IEEEauthorrefmark{3}Department of Electrical Engineering and Automation, Aalto University, Finland\\ \IEEEauthorrefmark{4} Department of Mechanical Engineering, Aalto University, Finland}
E-mail: \IEEEauthorrefmark{1}\IEEEauthorrefmark{2}\{zhao.dafang@ist,  chenz@sanken, i-tanigu@ist\}.osaka-u.ac.jp\\
\IEEEauthorrefmark{3}\IEEEauthorrefmark{4}\{zhengmao.li, xiaolei.yuan\}@aalto.fi}

\maketitle

\begin{abstract}
Heat, Ventilation and Air Conditioning (HVAC) systems play a critical role in maintaining a comfortable thermal environment and cost approximately 40\% of primary energy usage in the building sector. 
For smart energy management in buildings, usage patterns and their resulting profiles allow the improvement of control systems with prediction capabilities.
However, for large-scale HVAC system management, it is difficult to construct a detailed model for each subsystem.
In this paper, a new data-driven room temperature prediction model is proposed based on the $k$-means clustering method.
The proposed data-driven temperature prediction approach extracts the system operation feature through historical data analysis and further simplifies the system-level model to improve generalization and computational efficiency.
We evaluate the proposed approach in the real world.
The results demonstrated that our approach can significantly reduce modeling time without reducing prediction accuracy.
\end{abstract}


\section{Introduction}\label{sec:introduction}
% As one sustainable development goal is to reduce their energy consumption and associated carbon emissions, one area that remains to be optimized is indoor heating and cooling. 
One fundamental objective of sustainable development is to reduce energy consumption and associated carbon emissions.
Indoor heating and cooling remain a main area of focus and a challenge for further optimization.
% In fact, HVAC – which stands for Heating, Ventilation, and Air Conditioning – represents, on average, about 40\% of a building’s total energy use. 
In daily life, the heating, ventilation, and air conditioning (HVAC) system has a significant proportion of the energy use in a building, on average, about 40\% typically~\cite{21.500.11822_34572}.
% Methods that conserve electricity while still providing a comfortable indoor environment for workers could make a significant difference in the fight against climate change.
% {\color{magenta}
Therefore, effective methods that conserve electricity while still providing a comfortable indoor environment for workers are urged and might be potential for fighting against climate change.

% There is a great need for buildings to become more energy efficient and to be able to maintain interior comfort, reduce energy costs, and improve their environmental impact.
% The highest energy consumption in buildings is due to HVAC systems.
% Recent studies emphasize that efficient control on HVAC systems can significantly improve energy efficiency in buildings.
% Notably, suboptimal HVAC operations can precipitate disproportionate energy consumption. 
% However, in the practice of multi-HVAC systems management, building managers usually face the problem of the complexity of efficiently controlling and coordinating multi-HVAC systems to maintain optimal indoor environmental conditions.
% This includes challenges in energy consumption, system performance, maintenance, and integration of diverse equipment types.
% Achieving a balance between thermal comfort, energy efficiency, and sustainability is crucial for effective multi-HVAC systems.
Recently, numerous studies have been proposed on the topic of HVAC management.
Such methods are supported not only by simulations \cite{Kou2021,Li2021,Du2021}, but also by real-world implementation\cite{Parisio2014,Bursill2020,Maddalena2022}.
Experimental results demonstrate that optimal HVAC control can significantly reduce building energy consumption.
As a fundamental component of HVAC management, room temperature prediction enables HVAC systems to operate proactively, efficiently, and in a way that prioritizes occupant comfort while minimizing energy consumption and operational costs. 
% \Chen{Lack a summary: However, there is a drawback that is xxxxxx.}
Here, conventional solutions often involve physical and analytical models to calculate temperature changes.
However, this calculation estimates thermal parameters using parameter identification techniques with high computational costs. 
These modeling processes and their parameter estimations are manually and meticulously designed which requires detailed building information.
% Therefore, a reliable large-scale HVAC system should have the capacity for data fits that can automatically decide the scenario-oriented model.
% Although these models can provide detailed information on building thermal behavior and energy usage, the parameters are difficult to obtain or sometimes unavailable due to a variety of factors, including the lack of measurements, insufficient data, changes in building use, and limited accessibility.
This paper investigates a research question of ``\emph{how to build a reliable large-scale HVAC system should have the capacity for data fits that can automatically decide the scenario-oriented model.}''

With the rapid development of machine learning, they have been widely used in modeling building thermodynamics (i.e., temperature prediction).
Successful methods can be grouped into two categories:
i) rolling window that expands over time; and 
ii) fixed window that shifts over time.
The first category typically aims to forecast room temperature modeling long-term temporal dependencies, resulting in training with all available historical data.
For example, Furkan et al.~\cite{ELMAZ2021108327} uses Convolutional Neural Networks-Long Short Term Memory (CNN-LSTM) to conduct the prediction, achieving a high level of accuracy with an R$^2$ value greater than 0.9 in a 120-minute prediction horizon.
% However, the difference between the ideal state and the same attention given to different input data during model training will lead to prediction deviations.
Ben et al.~\cite{JIANG2022109536} propose an attention-LSTM architecture and show an advantage in both short-term and long-term temperature prediction. 
However, training with all historical data points (or time steps) inevitably brings redundant computational cost, including massive noise, which may decrease the forecasting performance.
% For instance, they might initially predict the next day temperature using only one day's worth of data. 
% Then, they might predict the subsequent day's temperature using the data from the just-predicted day, effectively using two days' worth of data.
% some studies believe room temperature only effected by short-term factors and not rely on long-term dependencies.
On the other hand, some studies solely utilize recent temporal room temperature variations for forecasting while avoiding redundant computation and contamination. 
Namely, Zhao et al.~\cite{Zhao2023} introduces fixed windows-based approach that uses symbolic regression (SR) to predict room temperature.
Despite its success, such a method is computation-intensive and relies on iteratively re-training processes for each forecasting step, such as every day. 

Based on the above discussion, we conjecture an accurate and reliable temperature prediction tailed for large-scale HVAC systems needs to model long-term dependencies while also giving attention to the correlation of the immediate previous time step. 
Therefore, our proposal is designed on top of the rolling and fixed window methods. 
More importantly, we introduce an efficient training strategy that utilizes clustering to mitigate the issues faced by prior works, i.e., the computational redundancy and associated costs.
% efficient HVAC system requires both modeling the long-term dependencies and paying attention to the correlation of previous single time steps.
% Hence, our proposal is designed on top of rolling window and fixed windows, more importantly, we introduce an efficient training strategy by clustering to further avoid the issues of previous works, i.e., computational redundancy and cost.  
% \Zhao{The authors believe that efficient temperature prediction models required for reliable large-scale HVAC systems should both consider long/short-term dependencies and avoid repeated training and contamination.}
%Based on the above discussion, how to merge the advantages of rolling window and fixed window-based approaches to achieve a lightweight and accurate temperature prediction model is a challenging problem.
% (Technical contributions)
% Therefore, it is necessary to develop a generic, lightweight temperature prediction model for more efficient HVAC management.
% redundant modeling and considerable computational overhead.
% predict the subsequent temperature only based on previous one time step. 
% In the category one, the research aims to establish an accurate and reliable room temperature prediction model. Paper 1, 2, 3.
% However, they suffer some drawbacks: xxxx.
% On the other hand, some studies pay attention to fixed xxx, namely, zhao, xxxx.
% Despite their success, xxxxxx (limitions)
% Based on these observations/discussion, our propsal is top of both modeling xxxxx.
% More importantly, clustering xxxxxx.
Several studies have shown that variation in room temperature and the operating patterns of HVAC systems manifests distinct intrinsic characteristics within the temporal domain~\cite{a16050256,FANG2021111053}.
These similarities can be harnessed to delineate specific operational scenarios of HVAC systems and further simplify the modeling process.
In this study, we propose a system scenario-based data-driven method to forecast the indoor temperature incorporated to a clustering machine learning approach.
We first extract characteristics of different time steps and use $k$-means clustering to classify and summarize system scenarios based on these characteristics. 
Instead of modeling all temporal time steps, each clustered scenario is represented using its relevant long-term and short-term temporal dependencies. This approach not only reduces computational costs but also prevents contamination from irrelevant information.
Furthermore, for online temperature forecasting, the proposed method will assign online data to the most appropriate system scenario based on its characteristics. This allows the online data to directly fit the pre-trained model instead of re-training. 
%Our system is potential for online-training. 

% This method adeptly leverages operational characteristics of HVAC systems, resulting in a marked reduction in computational overhead and the requisite number of models for multi-HVAC management.

% The rest of this paper is organized as follows. 
% Section~\ref{sec:prop-temp-pred} overviews our proposed system-scenario-based data-driven temperature prediction method;
% Section~\ref{sec:simu} presents the simulation results including comparisons with related studies.
% Finally, we conclude and describe directions for future work in Section~\ref{sec:concl}.

\section{Scenario-based Data-driven Temperature Prediction}\label{sec:prop-temp-pred}
% \subsection{Overview}
\begin{figure}[!t]
    \centering
    \includegraphics[width=0.7\linewidth]{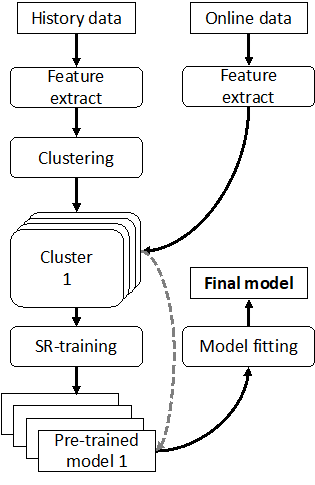}
    \caption{Diagram of temperature prediction workflow.}
    \label{fig:work-flow}
\end{figure}
The workflow of our proposal is illustrated in Figure~\ref{fig:work-flow}. 
The proposed method consists of the following steps:
\begin{enumerate}[label=(\alph*)]
    \item Collect historical sensing data and extract data feature.
    \item Classify the collected time series data according to the system scenario.
    \item Training temperature prediction model with symbolic regression for each cluster.
    \item Assign online data to existing cluster and selected pre-trained model.
    \item Fit the model coefficient with online data and generate a one-day ahead room temperature prediction model.
\end{enumerate}

\begin{figure}[!t]
    \centering
    \includegraphics[width=\linewidth]{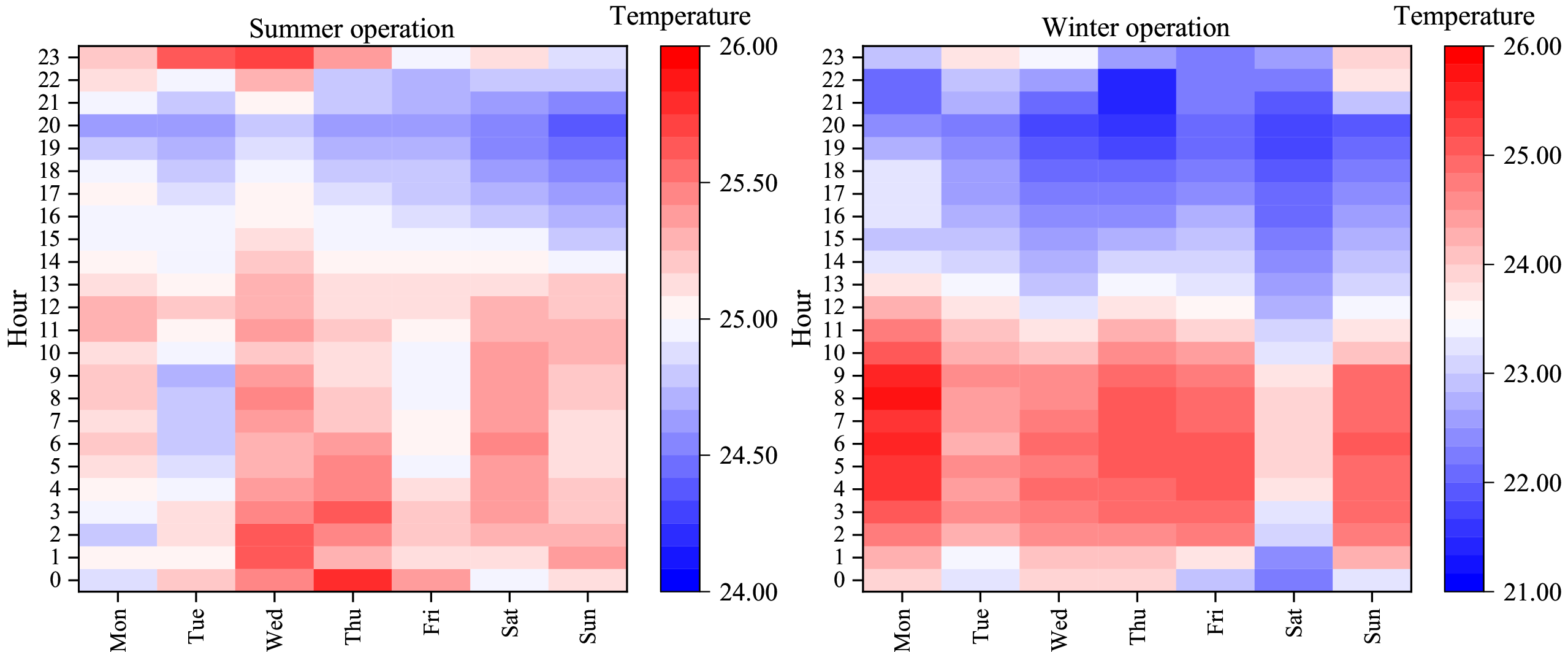}
    \caption{Room temperature distribution over week and day, along with the seasons.}
    \label{fig:temp-heatmap}
\end{figure}

\subsection{Data Collection and Preparation}
The data collection was done in the lecture building in Suita, Osaka, Japan. We selected one laboratory that is equivalent
to a typical office space. Its floor area is approximately 179 m$^2$; its ceiling is about 2.9 m high. The room has two parts: a 164 m$^2$ workspace and a 15 m$^2$ server area. The workspace contains 37 computers, including 20 GPU workstations. It has two exterior walls facing the south and east, easily affected by direct solar irradiation.
Five HVAC built-in indoor temperature sensors were placed on the roofs, one built-in outdoor temperature sensor and one power meter were placed on the rooftop HVAC outdoor unit.
The indoor/outdoor temperature and HVAC power data were collected at 5-minute intervals between July 10, 2022, and July 24, 2023.
During that period, over 237,576 data points were collected to evaluate the building thermodynamic.
A sample of the sensor data set is shown in Table \ref{tab:data-sample}.

Fig.~\ref{fig:temp-heatmap} shows the room temperature distribution over week and day. 
There is a strong difference between days of week during the summer (July-September) and winter (December to March).
It should be noticed that the early morning on Monday and Saturday in winter shows significant difference with other days which a transition effect is observed.
Therefore, by analyzing the time-series room temperature, HVAC power data, the hidden features can be used to classify different HVAC system operation scenario.

\begin{table}[!tb]
    \centering
        \caption{Sample of sensor data set}
    \label{tab:data-sample}
    \begin{tabular}{cccc}
    \toprule

         Time&  \begin{tabular}[c]{@{}c@{}}Indoor temp.\\ {[}$^\circ$C{]}\end{tabular} &  \begin{tabular}[c]{@{}c@{}}Outdoor temp.\\ {[}$^\circ$C{]}\end{tabular}& \begin{tabular}[c]{@{}c@{}}HVAC power\\ {[}W{]}\end{tabular}\\
    \midrule
         7/10/2022 08:00&  31&  30.8& 811.8
\\
         7/10/2022 08:05&  30&  30.8 & 0
\\
         7/10/2022 08:10&  30&  30.8& 0
\\
    \bottomrule
    \end{tabular}
\end{table}

\subsection{System Scenario Based Clustering}
The system scenario-based methodology is a generic and systematic design-time / run-time methodology to handle the dynamic nature of modern embedded systems.
It is based on the concept of \textit{System Scenarios}, which group system behaviors that are similar from a multidimensional cost perspective - such as resource requirements, delay, and energy consumption- in such a way that the system can be configured to exploit this cost similarity while reducing the overheads~\cite{Rao2012}.
% It combines a design-time phase, where these scenarios are individually optimized with a run-time resource manager that detects the appropriate scenario at run-time and switches between scenarios when necessary.
% There are several studies using scenario-based methodology in HVAC management~\cite{Parisio2014,6654024,theis2018}, the focusing point including but not limited to indoor air quality (IAQ), thermal comfort, electricity cost and etc.

\subsubsection{Time-series Data Feature Extraction}
Time series clustering has been shown effective in providing useful information in various applications\cite{tiano2021}. 
In this study, we adopted feature-based clustering method to classify HVAC system operation scenario.
Feature extraction is essential for turning the raw time-series data into a set of variables that capture the underlying patterns and structures in the data.
The statistical and frequency features are extracted for further clustering process.
These features including maximum, minimum, mean, median, variance, standard deviation, amplitude, real and imaginary parts, phase and phase offset of the sinusoids.
\subsubsection{System Scenario Clustering}
\begin{figure}
    \centering
    \begin{subfigure}[b]{\linewidth}
        \centering
        \includegraphics[width=0.8\linewidth]{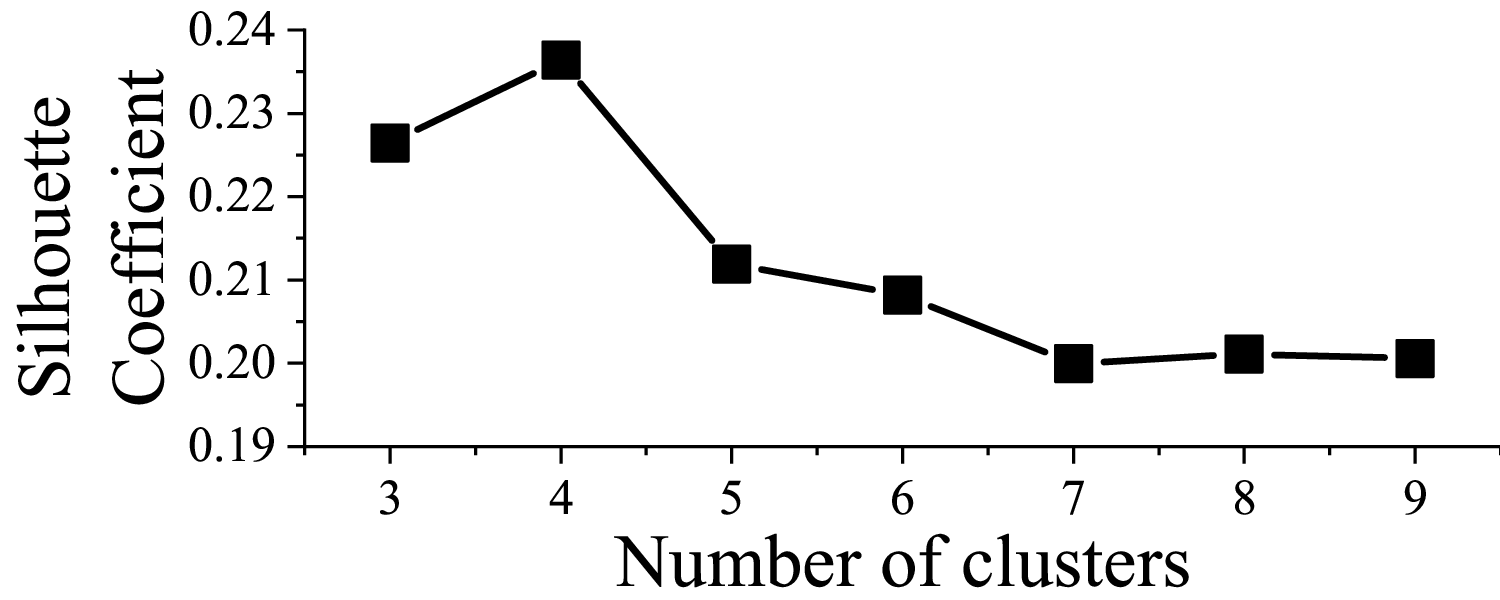}
        \caption{}
        % \caption{Silhouette coefficient for number of clusters.}
        \label{fig:num-cluster}
    \end{subfigure}
    \hfill
    \begin{subfigure}[b]{\linewidth}
        \centering
        \includegraphics[width=0.8\linewidth]{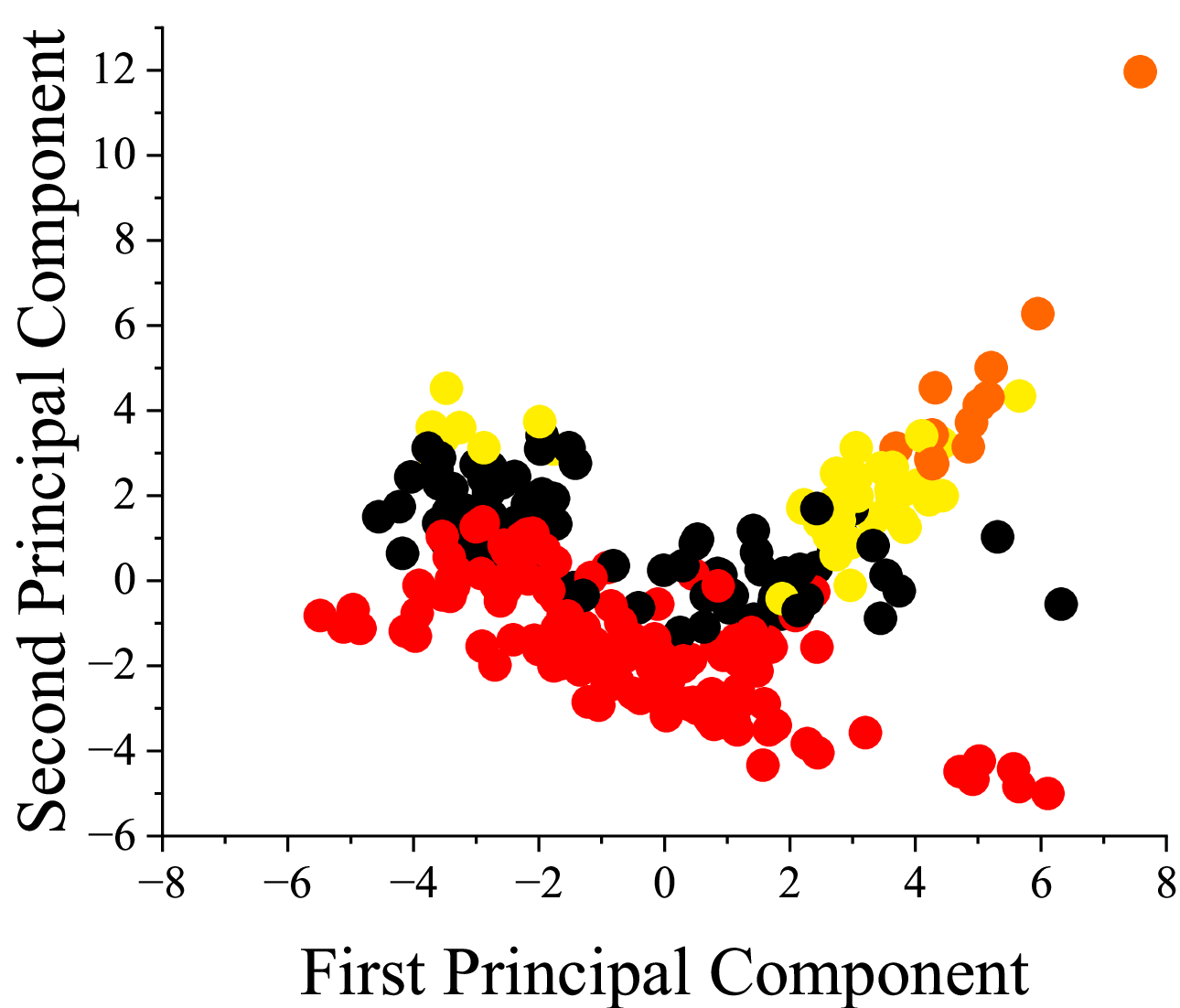}
        \caption{}
        % \caption{Clustering results for all observation period.}
        \label{fig:cluster-result}
    \end{subfigure}
    \caption{(a) Silhouette coefficient for number of clusters, (b) Clustering results for all observation period.}
\end{figure}

The main objective of clustering is to group similar data points together and discover the underlying system scenario.
The $k$-means method is used to clustering the time-series HVAC operation data.
To achieve this goal, $k$-means requires a fixed number of clusters.
This target cluster number, is referring to the number of centroids, which is an imaginary or real location of the center of the cluster. Then every data point is allocated to the nearest cluster while minimizing the intracluster variation.
The \textit{Silhouette Coefficient} metric is used to determine the optimal number of clusters.
Fig.~\ref{fig:num-cluster} shows the result of optimal number of clusters, it should be noticed that, the room temperature observation demonstrated that at least three scenarios exist in the target room. 
It shows that four clusters has maximum silhouette coefficient.
Hence, the optimal number of cluster is four in this study.
In the testing period, the temporal grand truth of clusters are labeled with the days of week, thus, seven clusters are classified.
Fig.~\ref{fig:cluster-result} shows the clustering results using $k$-means method. 
Moreover, the clustering evaluation metrics, Calinski-Harabasz Index, Davies-Bouldin Index, and Silhouette Score are 377.38, 0.42, and 0.69, respectively.
These metrics suggesting that the clusters have a relatively high degree of separation, relatively well-separated and distinct.
% \begin{table}
%     \centering
%         \caption{Result of Clustering Evaluation}
%     \begin{tabular}{cc}
%     \toprule
%          Metric& Value\\
%     \midrule
%          Calinski-Harabasz Index& 239.64\\
%          Davies-Bouldin Index& 0.44\\
%          Silhouette Score& 0.59\\
%     \bottomrule
%     \end{tabular}
%     \label{tab:clustering_metric}
% \end{table}

% \begin{figure}[!t]
%     \centering
%     \includegraphics[width=0.8\linewidth]{fig/Graph4-new.eps}
%     \caption{Silhouette coefficient for number of clusters.}
%     \label{fig:num-cluster}
% \end{figure}

% \begin{figure}[!t]
%     \centering
%     \includegraphics[width=0.8\linewidth]{fig/all_data_cluster.eps}
%     \caption{Clustering results for all observation period.}
%     \label{fig:cluster-result}
% \end{figure}

\subsection{Temperature Prediction with Symbolic Regression}
Even with the data-driven techniques evolution, barriers have arisen when it comes to the interpretation of the underlying physics that has somehow to be extracted from data patterns.
In this paper, an interpretable machine learning model, symbolic regression (SR), is adopted for one-day ahead prediction of room temperature.
SR is well-suited for discovering complex, nonlinear relationships in the data.
One advantage of SR is that it provides interpretable equations that can offer insights into the underlying relationships from data.
\begin{figure}[!t]
    \centering
    \includegraphics[width=0.8\linewidth]{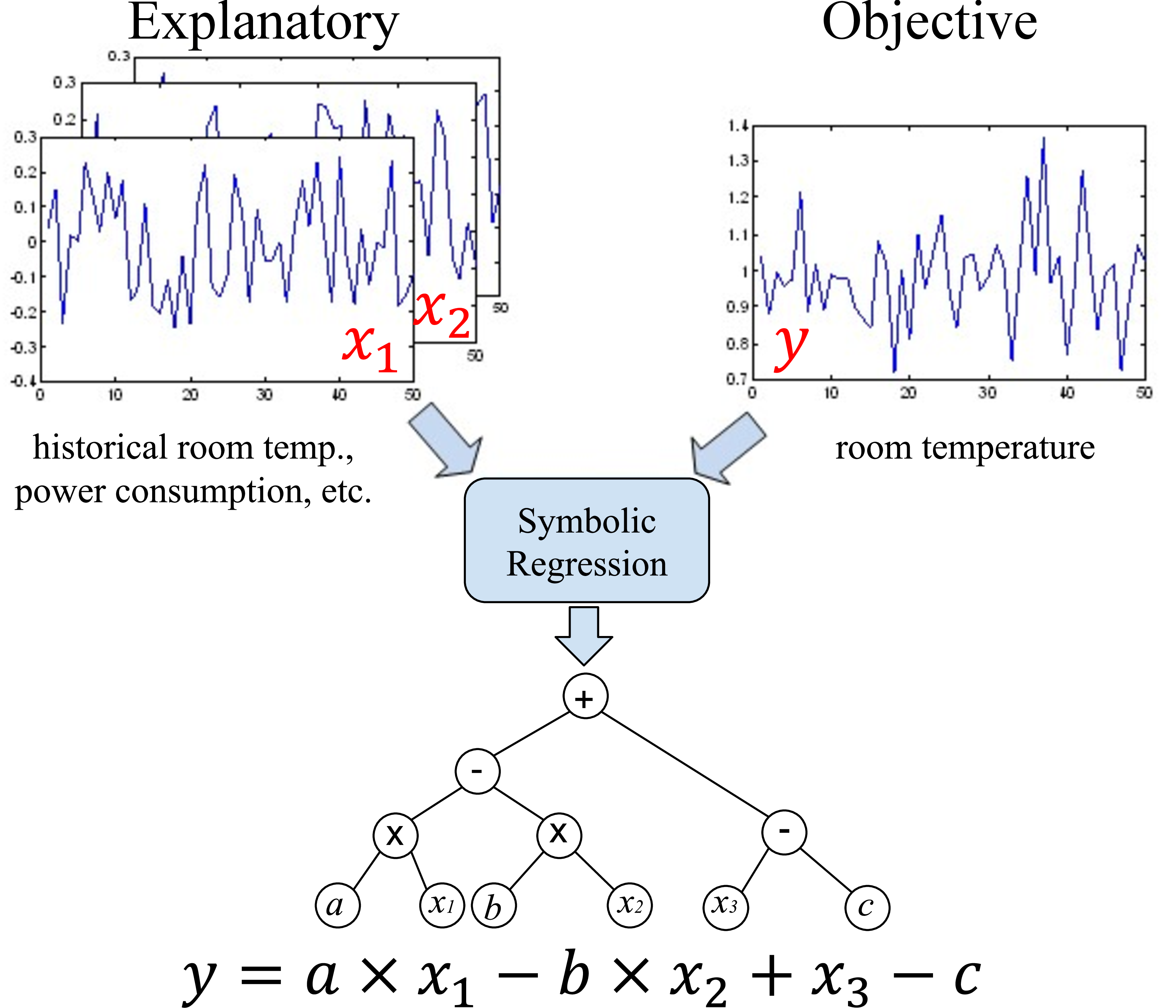}
    \caption{symbolic regression-based on a binary expression tree.}
    \label{fig:sr-diagram}
\end{figure}
Fig.~\ref{fig:sr-diagram} intuitively represents an expression tree structure in symbolic regression to find the mathematical expression that best fits a given dataset. In symbolic regression, a model is represented as a treelike structure, where nodes represent mathematical operations or functions applied to input explanatory variables, and leaves represent the final output or result of regression equations. The process of identifying the best expression involves adjusting the model parameters, such as the operations and variables used, to minimize the differences between the actual outputs and those predicted by the model. 

In this study, we extend the SR training process of \cite{Zhao2023}, where symbolic regression defines the combination $\bm{X}$ of input variables, such as room temperature $T^{in}$, outdoor temperature $T^{out}$, and HVAC power consumption $P$, as expressed as follows:
\begin{equation}
\bm{X} =
\begin{Bmatrix}
 T^{in},
 T^{out},
 P 
\end{Bmatrix}^{\intercal}.
\end{equation}

\begin{figure}[!t]
    \centering
    \includegraphics[width=0.8\linewidth]{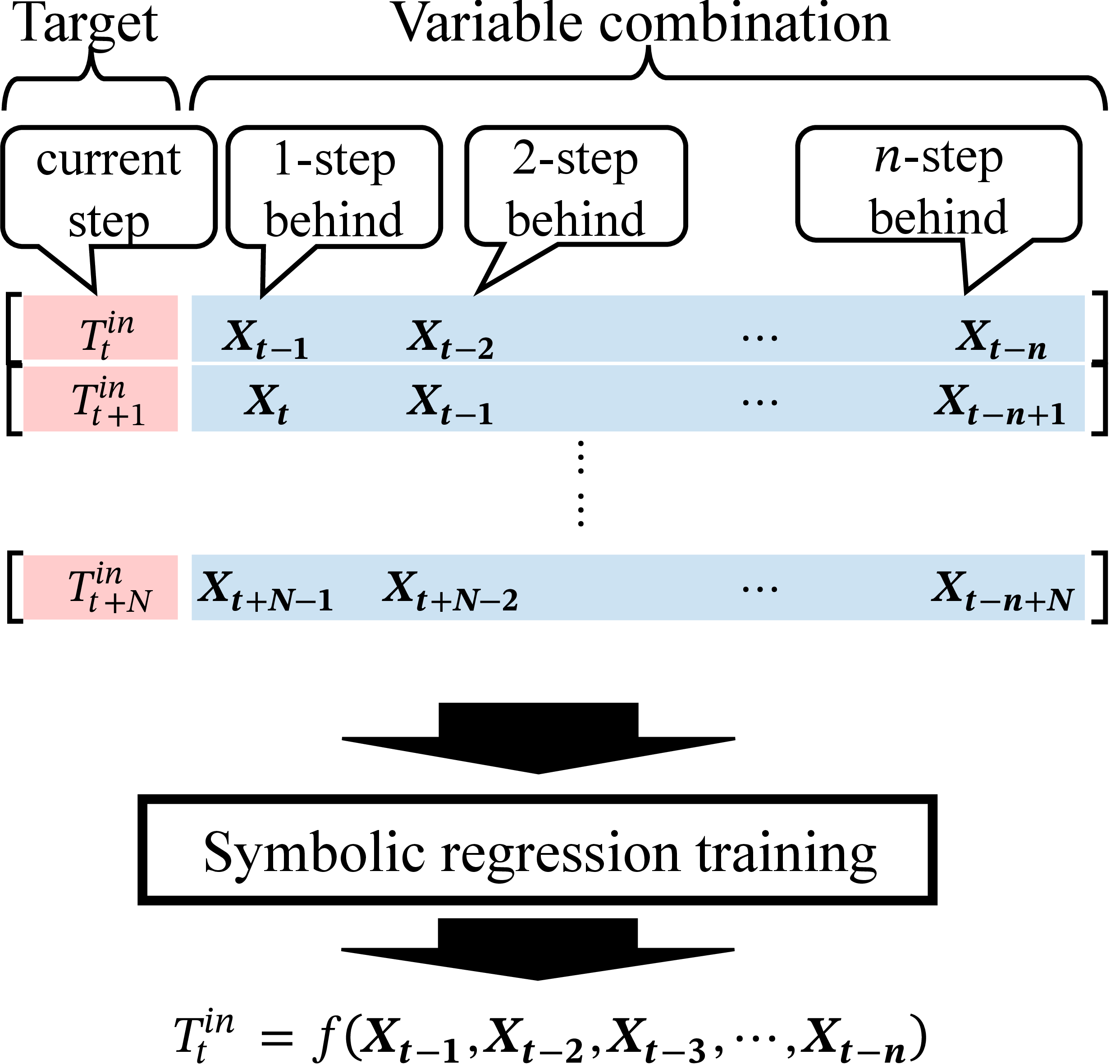}
    \caption{Overview of creating temperature prediction function using symbolic regression.}
    \label{fig:sr-overview}
\end{figure}

Different from previous research in this study we use pre-clustered data to training the SR model.
Fig.~\ref{fig:sr-overview} overviews the creation for temperature prediction function using symbolic regression.
Here for $N$ length of pre-clustered observation data are given as training data for symbolic regression. 
The current step is defined by the first time step in training data, where the prediction target for SR in training data are from current step to $N$ step ahead with one time step interval. 
The variable combination as training inputs are from one-step to $n$-step behind current step with one time step interval.
Predicted temperature $T^{in}_{t}$ for time step $t$ can be expressed by following function:
\begin{equation} \label{eq:sr_eq}
    T_{t}^{in} = f(\bm{X}_{t-1},\bm{X}_{t-2},\bm{X}_{t-3},\cdots,\bm{X}_{t-n}).
\end{equation}

In this study, the symbolic regression was performed using the PySR library~\cite{pysr}. 
Moreover, in the preliminary experiment we found that the training input variable combination $\bm{X_t}$ for past one-day can achieve the optimal balancing between computional cost and accuracy.
Therefore, we only choose past one-day data as training inputs, that is, $\bm{X_{t-1}}$, $\bm{X_{t-2}}$, $\dots$, $\bm{X_{t-288}}$, when the interval is 5 minutes.

\section{Simulation}\label{sec:simu}
To evaluate the data-driven scenario-based temperature prediction model performance, we conducted a simulation experiment and compared it with state-of-the-art (\textit{SOTA}) models~\cite{ELMAZ2021108327}.

In the simulation, we used the room temperature data obtained from the HVAC indoor units intake temperature sensor, the HVAC power data obtained from the power meter, and the outdoor temperature obtained from the HVAC outdoor units intake temperature sensor.
Since room thermodynamics differ significantly from season to season, we separated temperature prediction for HVAC cooling and heating operation.
For temperature prediction under cooling operation, we select training data from July to September, and from November to January for heating operation.
The proposed model is performed on a desktop\footnote{Dell OptiPlex 7070, Intel i7-9700 with 32GB RAM.} with sklearn and PySR library~\cite{pysr}.

\subsection{Temperature Prediction Under Cooling Operation}
\begin{figure}[!tbp]
\centerline{\includegraphics[width=\linewidth]{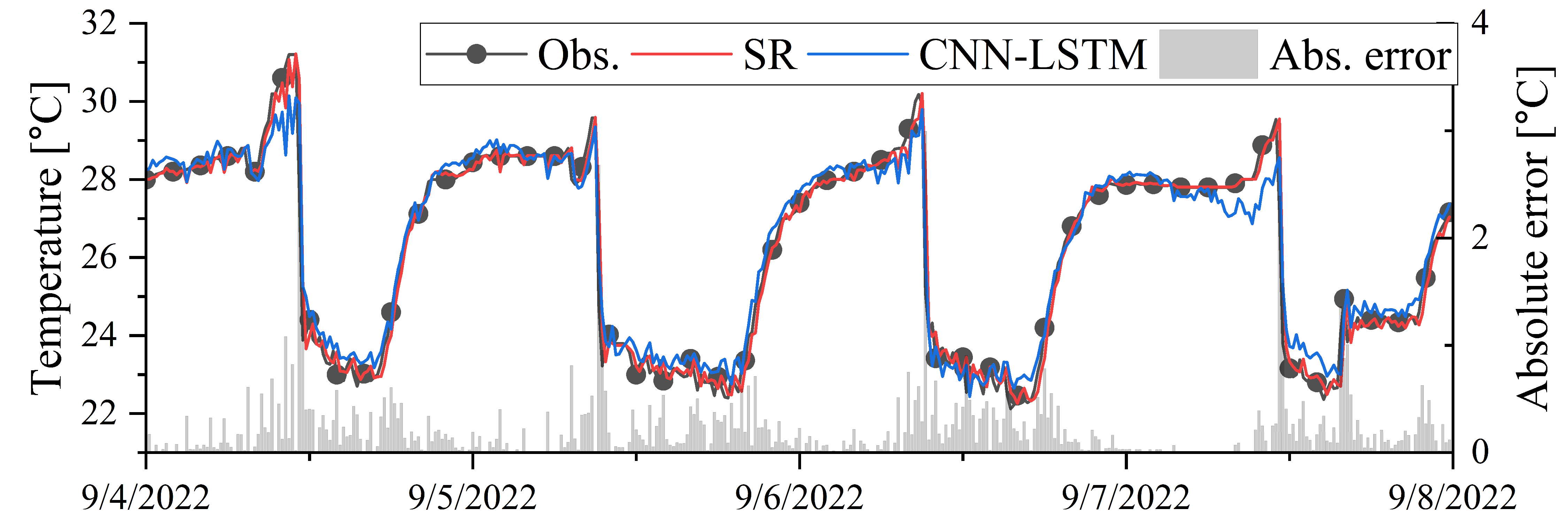}}
\caption[]{\label{fig:temp-pred-summer} Temperature prediction result under cooling operation.}
\end{figure}

For temperature prediction under cooling operation, the time-series room thermodynamics data have first been classified into four clusters with feature-based $k$-means clustering method.
% Fig.~\ref{fig:cluster-summer} shows the clustering results with three clusters.
% \begin{figure}[!tbp]
% \centerline{\includegraphics[width=0.7\linewidth]{fig/Graph14.eps}}
% \caption[]{\label{fig:cluster-summer} Clustering result for cooling operation.}
% \end{figure}
The first cluster has been used to evaluate the accuracy of the proposed prediction model in cooling operation.
This cluster contains 19 days of data, which is equivalent to 5403 data points.
Then split the variable combination $\bm{X}$ and the prediction target from first cluster into training and test sets with the ratio of 80:20.
The symbolic regression generated temperature prediction function is shown in Eq.\ref{eq:sr_eq_summer}.
\begin{align}
    \label{eq:sr_eq_summer}
    T^{in}_t =& 0.000073(T^{in}_{t-3}-23.49) \cdot \\ \nonumber &(T^{in}_{t-3}-P_{t-3}+0.2614 \cdot P_{t-72}) + T^{in}_{t-3}.
    % T^{in}_t &= -0.000150125(0.183209 \times P_{t-1}+2.18248)\\ \nonumber &\times (T^{in}_{t-1} - T^{in}_{t-84}) + T^{in}_{t-1}+0.0689856.
\end{align}
Fig.~\ref{fig:temp-pred-summer} shows the result of temperature prediction under cooling operation, where grey bar represent the absoulte error between observation and SR prediction, the black symbol line, red and blue line represent the observation, SR and CNN-LSTM prediction result, respectively.
We confirmed that the proposed model can predict the room temperature with high accuracy in cooling operation.
However, the prediction result shows that the proposed model has a tendency to slightly overestimate the room temperature.
% , and underestimates the room temperature when 
On the other hand, the CNN-LSTM model, shown by blue line, presents a tendency to underestimate the room temperature, especially in the early morning.
The CNN-LSTM model also shows a tendency to overestimate the room temperature in the night to midnight when HVAC stop cooling and building envelope releases heat cause temperature rising.

\subsection{Temperature Prediction Under Heating Operation}
\begin{figure}[!tbp]
\centerline{\includegraphics[width=\linewidth]{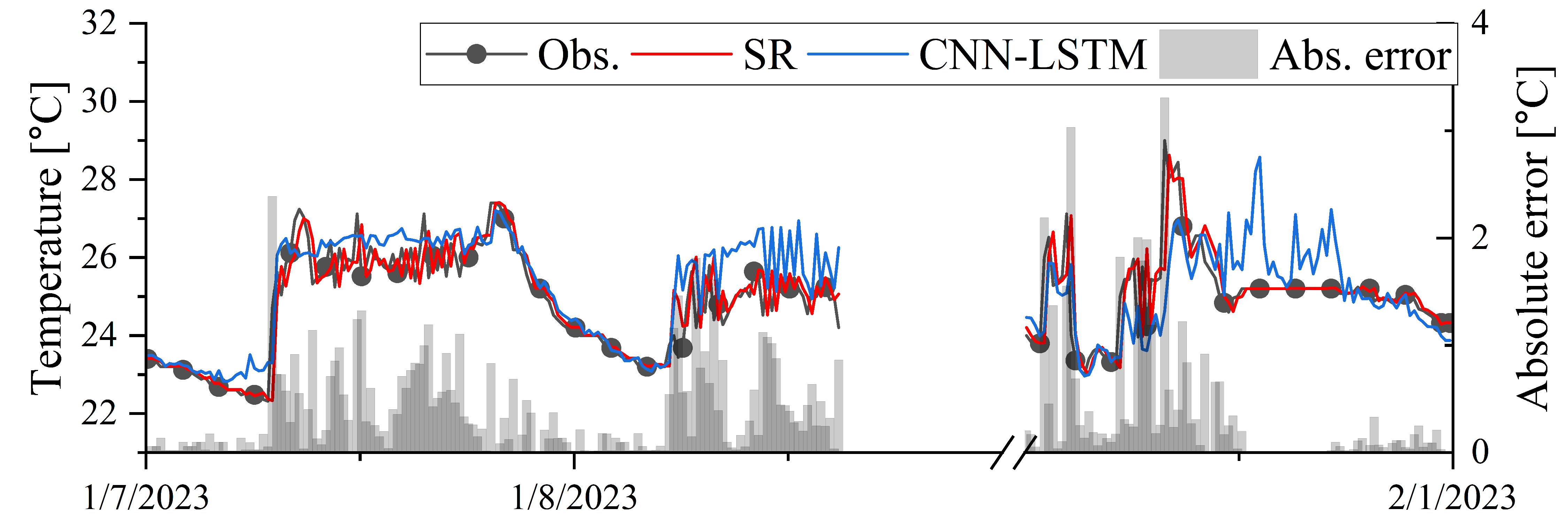}}
\caption[]{\label{fig:temp-pred-winter} Temperature prediction result under heating operation.}
\end{figure}

For temperature prediction under heating operation, the time-series room thermodynamics data have first been classified into five clusters with feature-based $k$-means clustering method.
% Fig.~\ref{fig:cluster-winter} shows the clustering results with three clusters.
% \begin{figure}[!tbp]
% \centerline{\includegraphics[width=0.7\linewidth]{fig/Graph13.eps}}
% \caption[]{\label{fig:cluster-winter} Clustering result for heating operation.}
% \end{figure}
Same to cooling operation, the first cluster has been used to evaluate the accuracy of the proposed prediction model in heating operation.
This cluster consists 11 days which contains to 2161 data points.
Then split the variable combination $\bm{X}$ and the prediction target from the first cluster into training and test sets with the ratio of 80:20.
The symbolic regression generated temperature prediction function is shown in Eq.\ref{eq:sr_eq_winter}.

\begin{align}
    \label{eq:sr_eq_winter}
    T^{in}_t = &T^{in}_{t-3}-0.00014(T^{in}_{t-3}-T^{in}_{t-103}-2.316) \cdot \\ \nonumber &(0.3573 \cdot (P_{t-3}-T^{out}_{t-36})+T^{in}_{t-72}).
    % 0.000073(T^{in}_{t-3}-23.49) \cdot \\ \nonumber &(T^{in}_{t-3}-P_{t-3}+0.2614 \cdot P_{t-72}) + T^{in}_{t-3}.
    % T^{in}_t &= -0.000150125(0.183209 \times P_{t-1}+2.18248)\\ \nonumber &\times (T^{in}_{t-1} - T^{in}_{t-84}) + T^{in}_{t-1}+0.0689856.
\end{align}
Fig.~\ref{fig:temp-pred-winter} shows the result of temperature prediction under heating operation,
where grey bar represent the absoulte error between observation and SR prediction, the black symbol line, red and blue line represent the observation, SR and CNN-LSTM prediction result, respectively.
We have confirmed that the proposed model can predict the room temperature with relatively high accuracy in heating operation.
Unlike cooling operation, the proposed model has a tendency to slightly underestimate the room temperature.
Moreover, the proposed model showing a slightly time lag compared with observation.
On the other hand, , the CNN-LSTM model, shown by blue line, presents a tendency to overestimate the room temperature.

Table~\ref{tab:temp-result} shows the comparison of the proposed model and the CNN-LSTM model (\textit{SOTA}).
The accuracy of the predictions made by the symbolic regression was also superior in all the above terms.
Moreover, the proposed model has a shorter training time than the CNN-LSTM model where the training can be reduced up to 33\%.

\begin{table}[]
    \centering    
    \caption{Result of temperature prediction for cooling and heating operation\label{tab:temp-result}}
    \begin{tabular}{lllll}
        \toprule
     & \multicolumn{2}{c}{Cooling} & \multicolumn{2}{c}{Heating} \\
     \midrule
     Metrics & SR          & \textit{SOTA}          & SR          & \textit{SOTA}          \\
     \midrule
     R$^2$&  0.972          &     0.962     & 0.753           &   0.571       \\
     RMSE [$^{\circ}$C]& 0.409            &  0.479       &   0.621        & 0.866         \\
     MAE&   0.218          &    0.343        &   0.393          &   0.379        \\
     MAPE [\%]&  0.864           &   0.013       &      1.495       &       2.392        \\
     MSE& 0.167        &     0.229    &        0.385     &       0.750    \\
     Training Time [s]& 70.2         &   101.1           & 52.4            & 79.3\\
     \bottomrule
    \end{tabular}
    \end{table}

% \begin{table}
%     \centering
%         \begin{tabular}{ccc}
%         \toprule
%              & \begin{tabular}[c]{@{}c@{}}Proposed\\ model\end{tabular} & \begin{tabular}[c]{@{}c@{}}SOTA\\ model\end{tabular} \\
%         \midrule
%                 MAE& 0.17& 0.48\\
%                 RMSE& 0.33& 0.61\\
%                 R$^2$& 0.97& 0.98\\
%                 MAPE& 0.62& 0.02\\
%                 MSE& 0.11& 0.37\\
%                 Training Time [sec] & 44.19& 91.23\\
%         \bottomrule
%         \end{tabular}
% \end{table}

\section{Conclusion}\label{sec:concl}
In this study, we proposed a system scenario-based data-driven room temperature prediction model for HVAC systems management.
The proposed method adeptly leverages operational characteristics of HVAC systems, resulting in a marked reduction in computational overhead and the requisite number of models for multi-HVAC management.
We evaluated the proposed approach with real world operation data.
The results demonstrated that our approach can significantly reduce modeling time without reducing prediction accuracy.
In the future, we will evaluate the proposed method in a real-world environment and further improve the prediction accuracy.

\bibliographystyle{IEEEtran}
\bibliography{ref.bib}

\end{document}